\documentclass[10pt,twocolumn,letterpaper]{article}

\usepackage{cvpr}              
\usepackage{graphicx}
\usepackage{booktabs}
\usepackage{multirow}
\usepackage{xcolor}
\usepackage{colortbl}
\usepackage{amsmath}

\usepackage[accsupp]{axessibility}  

\usepackage{graphicx}
\usepackage{tikz}
\tikzstyle{bag} = [align=center]
\usetikzlibrary{positioning, quotes, arrows.meta}
\usepackage{amssymb}

\definecolor{cvprblue}{rgb}{0.21,0.49,0.74}
\usepackage[pagebackref,breaklinks,colorlinks,allcolors=cvprblue]{hyperref}

\def\paperID{*****} 
\def\confName{CVPR}
\def\confYear{2025}

\title{Defending Against Frequency-Based Attacks with Diffusion Models}

\author{
    Fatemeh Amerehi \quad Patrick Healy \\  
    Computer Science and Information Systems \\  
    University of Limerick \\  
    {\tt\small Fatemeh.Amerehi@ul.ie, Patrick.Healy@ul.ie}  
}

\begin{document}
\maketitle
\begin{abstract}

Adversarial training is a common strategy for enhancing model robustness against adversarial attacks. However, it is typically tailored to the specific attack types it is trained on, limiting its ability to generalize to unseen threat models.
Adversarial purification offers an alternative by leveraging a generative model to remove perturbations before classification. Since the purifier is trained independently of both the classifier and the threat models, it is better equipped to handle previously unseen attack scenarios. Diffusion models have proven highly effective for noise purification, not only in countering pixel-wise adversarial perturbations but also in addressing non-adversarial data shifts.
In this study, we broaden the focus beyond pixel-wise robustness to explore the extent to which purification can mitigate both spectral and spatial adversarial attacks. Our findings highlight its effectiveness in handling diverse distortion patterns across low- to high-frequency regions.

\end{abstract}

\section{Introduction}
\label{sec:intro}

The accuracy of neural network predictions is often compromised by distributional shifts, whether adversarially crafted~\cite{biggio2013evasion, szegedy2013intriguing, goodfellow2014explaining} or naturally occurring~\cite{azulay2018deep, hendrycks2018benchmarking}, and whether these shifts manifest at the pixel level~\cite{madry2017towards, touvron2019fixing} or the frequency level~\cite{luo2022frequency, deng2020frequency}.

Pixel-wise adversarial attacks are designed to be minimally detectable, leveraging subtle perturbations to mislead models while preserving visual similarity~\cite{goodfellow2014explaining}. Their imperceptibility is typically measured using \( \ell_p \) norms~\cite{carlini2017towards, madry2017towards, kurakin2018adversarial, modas2019sparsefool, chen2023imperceptible} and, in some cases, human perception ~\cite{luo2018towards, zhao2020towards, shamsabadi2020colorfool, deng2020frequency}.

These perturbations are typically crafted through an optimization process to maximize the model’s loss and are then added to the original input~\cite{moosavi2016deepfool, kurakin2016adversarial, madry2017towards, kurakin2018adversarial}.
The Fast Gradient Sign Method (FGSM) ~\cite{goodfellow2014explaining} 
approximates this maximization by performing a single gradient step to identify the direction in which the loss increases most rapidly.  Iterative extensions, such as the Basic Iterative Method~\cite{kurakin2016adversarial} and Projected Gradient Descent (PGD)~\cite{madry2017towards}, strengthen FGSM by applying multiple iterations to refine the adversarial perturbation. AutoAttack \cite{croce2020reliable} further advances these techniques by combining two parameter-free PGD variants with the Fast Adaptive Boundary attack~\cite{croce2020minimally} and the Square Attack~\cite{andriushchenko2020square}, enabling both white-box and black-box attacks.

 Frequency-based attacks modify images by adjusting energy in specific frequency bands~\cite{luo2022frequency, deng2020frequency, kim2024exploring}. This often involves transforming the image into the frequency domain, making subtle modifications, and then converting it back, all while preserving visual quality and minimizing perceptible noise~\cite{sharma2019effectiveness, huang2021rda}.
AdvDrop~\cite{duan2021advdrop} removes features essential to the model yet unnoticeable to human observers by quantitatively reducing targeted frequency components~\cite{duan2021advdrop}. 

Frequency-domain filtering constrains perturbations to high-frequency components~\cite{luo2022frequency} or narrows the search for adversarial images to low-frequency regions~\cite{guo2018low}. Frequency-selective attacks independently manipulate amplitude and phase components, enabling more precise control over perturbations~\cite{kim2024exploring}. Some methods introduce large perturbation magnitudes~\cite{huang2021rda}, while others accelerate feature collapse in vision transformers~\cite{dong2021attention} by exploiting their reliance on low-frequency components~\cite{gao2024attacking}. 
F-mixup~\cite{li2021f} generates adversarial examples by blending the low-frequency component of one class with the high-frequency component of another. Frequency Dropout~\cite{zhu2024frequency} reduces certain gradient frequency coefficients to improve attack transferability, assuming that high-frequency components contain noise and overfitting details related to the substitute model. AdvINN~\cite{chen2023imperceptible} utilizes Discrete Wavelet Transform to decompose the clean and target images into low- and high-frequency components. It then generates adversarial samples by exchanging feature-level information, suppressing discriminative details in the clean images, and embedding class-specific attributes from the target samples.

Beyond these methods, other attack strategies employ generative models~\cite{jalal2017robust, chen2023advdiffuser}, alter pixel positions~\cite{xiao2018spatially}, or leverage transformations such as subtly modifying shape~\cite{xiao2018spatially}, as well as modifying texture and color~\cite{afifi2019else,Bhattad2020Unrestricted,shamsabadi2020colorfool}, or combining all of these within the image content~\cite{chen2024content}.

In response, various defense mechanisms exist, including defensive distillation~\cite{papernot2016distillation}, feature squeezing~\cite{xu2017feature}, logit pairing~\cite{kannan2018adversarial}, adversarial detection~\cite{metzen2017detecting, pang2018towards}, gradient regularization~\cite{tramer2017ensemble,wu2020adversarial}, adversarial training~\cite{goodfellow2014explaining, madry2017towards, wang2023better}, and adversarial purification~\cite{yoon2021adversarial, nie2022diffusion, lin2024adversarial, li2025adversarial}. 

Among these approaches, adversarial training has been shown to be highly effective~\cite{goodfellow2014explaining, kurakin2016adversarial, carlini2017adversarial_b, moosavi2016deepfool, athalye2018obfuscated, bai2021recent}. However, it often comes with a trade-off between robustness and accuracy~\cite{tsipras2018robustness, zhang2019theoretically}, as well as between in- and out-of-distribution generalization~\cite{zhang2019theoretically}. It can also interfere with a model’s invariance to input transformations~\cite{singla2021shift,kamath2021can}
, while amplifying performance disparities 
across different classes \cite{benz2021robustness, ma2022tradeoff}. 
Moreover, adversarial training shows limited generalization to unseen attacks~\cite{stutz2020confidence, laidlaw2021perceptual}. While models exhibit robustness against the specific threat model they were trained on, they often remain vulnerable to others~\cite{tramer2019adversarial}. For example, models trained with \(\ell_\infty\) threat models may lack robustness to \(\ell_2\) or spatial adversarial perturbations, and vice versa~\cite{nie2022diffusion}.

Unlike adversarial training, adversarial purification uses a generative model to remove adversarial perturbations from images prior to classification~\cite{yoon2021adversarial}.
This approach offers greater flexibility, as the purification models are trained independently without any assumptions about the attack type or classifier, making them effective against a wide range of previously unseen adversarial perturbations~\cite{nie2022diffusion}.
The use of diffusion models~\cite{ho2020denoising, song2021scorebased, dhariwal2021diffusion} for noise purification has proven highly effective in countering pixel-wise adversarial perturbations, consistently outperforming adversarial training in \(\ell_\infty\) and \(\ell_2\) threat models~\cite{nie2022diffusion}, as well as in certified adversarial robustness~\cite{carlini2023certified}. Moreover, their effectiveness extends beyond adversarial robustness to broader data shifts, including variations in style~\cite{yu2023distribution}, as well as non-adversarial perturbations~\cite{gao2023back}.  Moving beyond pixel-wise robustness, this paper broadens the lens and provides insight into how purification handles shifts in both spatial and spectral domains.

\section{Related Works }

Having explored adversarial attacks, the following section reviews literature on methods for addressing unseen attack scenarios and adversarial purification techniques.

\textbf{Robustness Against Unforeseen Attacks.}  Many defense mechanisms offer guarantees based on specific threat models~\cite{Croce2020Provable} or integrate knowledge of the threat model during training ~\cite{madry2017towards}. Adversarial training, while effective against known threats, primarily relies on generated perturbations and thus struggles to generalize to unseen attacks. It typically assumes the adversary follows a predefined threat model, such as \( \ell_p \)-bounded perturbations within a fixed budget, which may not always hold true~\cite{laidlaw2021perceptual}.

One approach to mitigating unseen threats involves generating adversarial examples for all types of perturbations and training on either all of these examples or focusing on the worst-case scenario~\cite{tramer2019adversarial}. Dual Manifold Adversarial Training leverages information about the underlying manifold of natural images, employing adversarial perturbations in both the latent and image spaces, enabling generalization to both \( \ell_p \)-bounded and non-norm-bounded threats. Variation Regularization reduces the variation in the feature extractor across the source threat model during training, improving the model's ability to generalize to previously unseen attacks~\cite{Dai2022Formulating}. Confidence-calibrated adversarial training introduces a target distribution that favors uniformity for large perturbations, allowing the model to extrapolate beyond the threat model employed during training ~\cite{stutz2020confidence}. 
Additionally, some methods offer bounds for generalizing to an unknown adversary with oracle access during training~\cite{montasser2021adversarially}.

\textbf{Adversarial purification.} Adversarial purification employs a standalone model that removes adversarial noise from potentially attacked images, restoring them to a clean state for classification~\cite{song2018pixeldefend, srinivasan2021robustifying, yoon2021adversarial,nie2022diffusion}.
Defense-GAN~\cite{samangouei2018defensegan} leverages Wasserstein GANs~\cite{arjovsky17a} to model the distribution of unperturbed images and, during inference, reconstructs an image free of adversarial perturbations by finding a latent vector that minimizes the difference between the generated and input images, which is then classified ~\cite{samangouei2018defensegan}.
PixelDefend~\cite{song2018pixeldefend} utilizes the autoregressive generative model PixelCNN~\cite{van2016conditional} to detect and purify adversarial examples.
Some approaches employ energy-based models~\cite{Grathwohl2020Your, du2019implicit, srinivasan2021robustifying} to purify attacked images through Langevin dynamics, while others rely on denoising score-based generative models~\cite{song2019generative,yoon2021adversarial}.
DiffPure~\cite{nie2022diffusion} uses diffusion models to purify adversarial examples prior to inputting them into classifiers. A guided diffusion model for adversarial purification employs the difference between an adversarial example and a purified one as a form of guidance~\cite{wang2022guided}. Alternatively, it may utilize an auxiliary neural network, trained through adversarial learning, to steer the reverse diffusion process based on latent representation distances rather than pixel-level values~\cite{lin2024adversarial}. Another approach involves the use of contrastive guidance~\cite{bai2024diffusion}. 

DensePure~\cite{xiao2023densepure} performs multiple denoising steps with different random seeds to generate reversed samples, which are then classified and combined through majority voting for the final prediction.
Other methods use random transforms to prevent overfitting to known attacks and fine-tune the purifier model with adversarial loss to enhance robustness~\cite{lin2024adversarial_pure}, or adopt a gradual noise-scheduling strategy to strengthen diffusion-based purification~\cite{lee2023robust}. While these works primarily investigate pixel-wise adversarial perturbations in the spatial domain,  we redirect the focus toward frequency-based perturbations in the spectral domain, as it remains a less-studied aspect of adversarial purification and robustness.

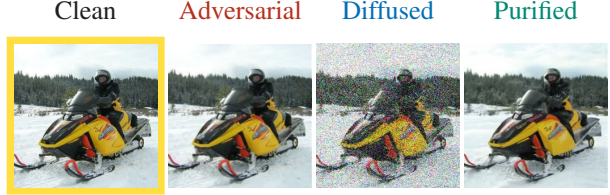
\begin{figure*}[h]
    \centering
    \begin{tikzpicture}

        \node[draw=Goldenrod, line width=2mm, inner sep=0pt] (img1) at (0,0) 
        {\includegraphics[width=1.9cm]{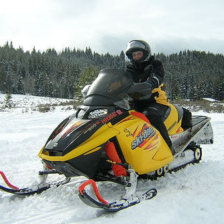}};
        \node[above=1mm, align=center] at (img1.north) {\textcolor{Black}{Clean}};
        \node[inner sep=0pt, right=0.05 of img1] (img2) {\includegraphics[width=1.9cm]{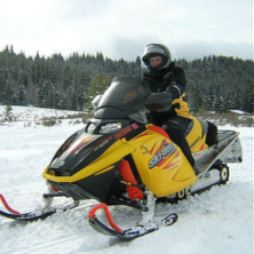}};
        \node[above=2mm, align=center] at (img2.north) {\textcolor{BrickRed}{Adversarial}};
        \node[inner sep=0pt, right=0.05 of img2] (img3) {\includegraphics[width=1.9cm]{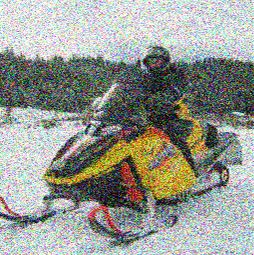}};
        \node[above=2mm, align=center] at (img3.north) {\textcolor{NavyBlue}{Diffused}};
        \node[inner sep=0pt, right=0.05 of img3] (img4) {\includegraphics[width=1.9cm]{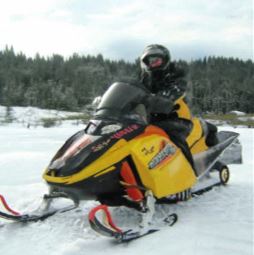}};
        \node[above=2mm, align=center] at (img4.north) {\textcolor{PineGreen}{Purified}};

    \end{tikzpicture}
    \caption{Clean, adversarial, diffused, and purified images. The clean image is the original, uncorrupted image from ImageNet. The adversarial image is generated for ResNet-50 by perturbing all components (magnitude, phase, and pixel values) of the image. The adversarial image is then purified using diffusion purification. The diffused image corresponds to \( t^* = 0.15 \), and the purified image is obtained at \( t = 0 \).}
    \label{fig:clean_adv_diff_pure}
\end{figure*}
\vspace{5pt}

\begin{figure*}
    \centering
    \begin{tikzpicture}
        \def\imagenames{
            {RN_phase.png, 
            RN_phase_purt.png,
            RN_phase_hist.png,
            RN_mag.png,
            RN_mag_purt.png,
            RN_mag_hist.png,
            RN_pixel.png,
            RN_pixel_purt.png,
            RN_pixel_hist.png},
            {ViT_phase.png,
            ViT_phase_purt.png,
            ViT_phase_hist.png,
            ViT_mag.png,
            ViT_mag_purt.png,
            ViT_mag_hist.png,
            ViT_pixel.png,
            ViT_pixel_purt.png,
            ViT_pixel_hist.png},
            {Swin_phase.png,
            Swin_phase_purt.png,
            Swin_phase_hist.png,
            Swin_mag.png,
            Swin_mag_purt.png,
            Swin_mag_hist.png,
            Swin_pixel.png,
            Swin_pixel_purt.png,
            Swin_pixel_hist.png},         
        }
        
        \foreach \row [count=\r] in \imagenames {
            \foreach \col [count=\c] in \row {
                \node[inner sep=0pt] (img\c-\r) at (\c*1.8, 4-\r*1.8) {\includegraphics[width=1.7cm]{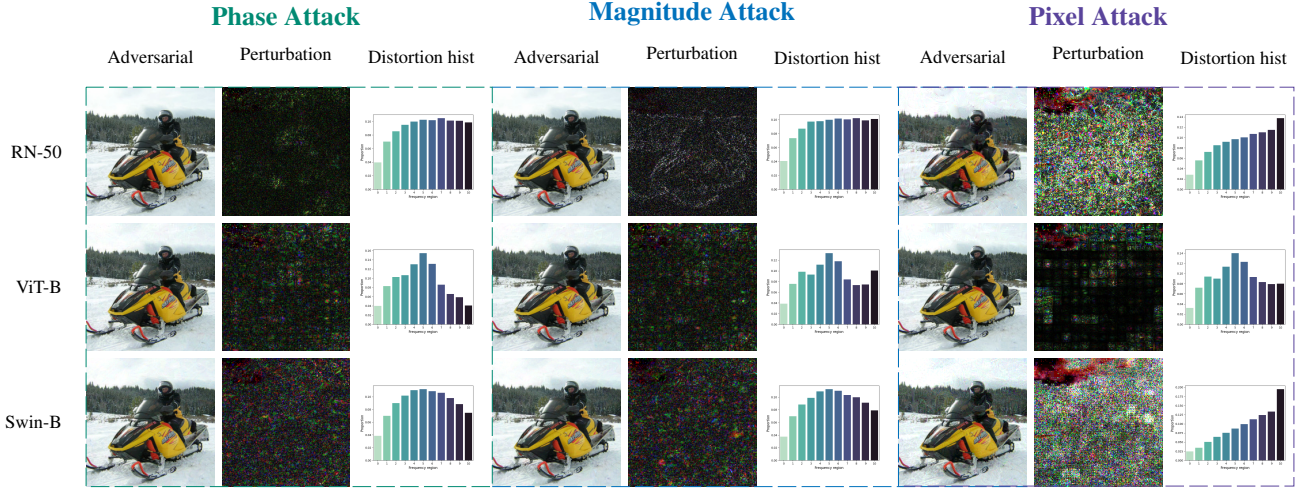}};
            }
        }
        \node[anchor=east, xshift=-2mm] at (img1-1.west) {\scriptsize RN-50};
        \node[anchor=east, xshift=-2mm] at (img1-2.west) {\scriptsize ViT-B};
        \node[anchor=east, xshift=-2mm] at (img1-3.west) {\scriptsize Swin-B};
        \node [above=7mm of img2-1.north] {\textcolor{PineGreen}{\textbf{Phase Attack}}};
        \node [above=2mm of img1-1.north] {\scriptsize Adversarial};
        \node [above=2.3mm of img2-1.north] {\scriptsize Perturbation};
        \node [above=4.1mm of img3-1.north] {\scriptsize Distortion hist};
        \node [above=7mm of img5-1.north] {\textcolor{RoyalBlue}{\textbf{Magnitude Attack}}};
        \node [above=2mm of img4-1.north] {\scriptsize Adversarial};
        \node [above=2.5mm of img5-1.north] {\scriptsize Perturbation};
        \node [above=4mm of img6-1.north] {\scriptsize Distortion hist};
        \node [above=7mm of img8-1.north] { \textcolor{Violet}{\textbf{Pixel Attack}}};
        \node [above=2mm of img7-1.north] {\scriptsize Adversarial};
        \node [above=2.5mm of img8-1.north] {\scriptsize Perturbation};
        \node [above=4mm of img9-1.north] {\scriptsize Distortion hist};
      \draw[ ultra thin, draw=PineGreen, dash pattern=on 7pt off 2pt](img1-1.north west)  rectangle (img4-3.south west) ;
      %
      \draw[ ultra thin, draw=RoyalBlue, dash pattern=on 7pt off 2pt](img4-1.north west) -- (img7-1.north east); 
      \draw[ ultra thin, draw=RoyalBlue, dash pattern=on 7pt off 2pt](img4-3.south west) -- (img7-3.south east); 
      \draw[ ultra thin, draw=Violet, dash pattern=on 7pt off 2pt](img7-1.north west)  rectangle (17,-2.25) ;
      \draw[ ultra thin, draw=RoyalBlue, dash pattern=on 7pt off 2pt](img7-1.north west) -- (img7-3.south west);

    \end{tikzpicture}
    \caption{Adversarial examples generated by perturbing the magnitude, phase, and pixel values across various architectures. The perturbations, representing the differences between the original and attacked images (magnified by a factor of 20 for visualization). The distortion histograms, obtained by applying the Fourier transform to the perturbations, highlight the impact of each attack on the spectral characteristics of the images. In ResNet-50~\citep{he2016deep}, the distortion is primarily concentrated in high-frequency regions, while ViT-B~\cite{dosovitskiy2021an} and Swin-B~\citep{liu2021swin} exhibit distortions mainly in the mid-to-low frequency ranges.}
    \label{fig:Frequency_adversarial_examples}
\end{figure*}


\begin{figure*}[h]
    \centering
    \begin{tikzpicture}

        \node[draw=Goldenrod, line width=2mm, inner sep=0pt] (img1) at (0,0) 
        {\includegraphics[width=1.9cm]{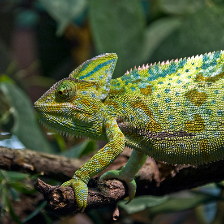}};
        \node[above=9mm, align=center] at (img1.north) {\textcolor{red}{Adversarial}};
        \node[above=5mm, align=center] at (img1.north) {\textcolor{red}{ (Magnitude/Phase/Pixel)}};
        \node[above=1mm] at (img1.north) {\textcolor{black}{$t=0$}};

        \node[inner sep=0pt, right=0.05 of img1] (img2) {\includegraphics[width=1.9cm]{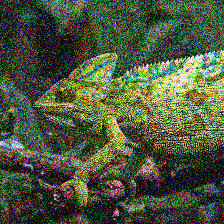}};
        \node[inner sep=0pt, right=0.05 of img2] (img3) {\includegraphics[width=1.9cm]{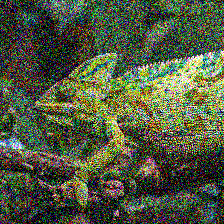}};
        \node[inner sep=0pt, right=0.05 of img3] (img4) {\includegraphics[width=1.9cm]{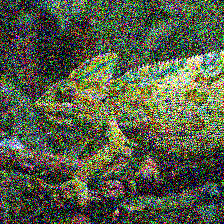}};
        \node[inner sep=0pt, right=0.05 of img4] (img5) {\includegraphics[width=1.9cm]{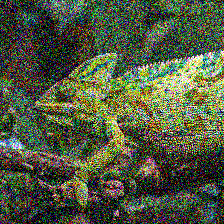}};
        \node[inner sep=0pt, right=0.05 of img5] (img6) {\includegraphics[width=1.9cm]{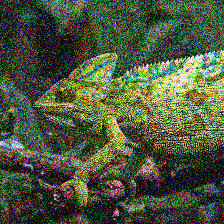}};

        \node[draw=SeaGreen, line width=2mm,, inner sep=0pt, right=0.05 of img6] (img7) 
        {\includegraphics[width=1.9cm]{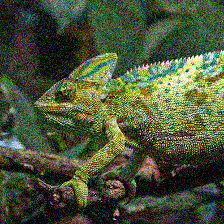}};

        \node[above=9mm] at (img7.north) {\textcolor{SeaGreen}{Purified}};
        \node[above=5mm] at (img7.north) {\textcolor{SeaGreen}{ chameleon}};
        \node[above=1mm] at (img7.north) {\textcolor{black}{$t=0$}};
        \node[above=7mm] at (img4.north) {\textcolor{black}{Diffused}};
        \node[above=1mm] at (img4.north) {\textcolor{black}{$t=t^*$}};

        \draw[-Triangle, thick] ([yshift=12mm]img2.west) -- ([yshift=12mm]img3.east) 
            node[midway, above] {\textcolor{black}{\scriptsize Forward Process}};

        \draw[-Triangle, thick] ([yshift=12mm]img5.west) -- ([yshift=12mm]img6.east) 
            node[midway, above] {\textcolor{black}{\scriptsize Reverse Process}};

    \end{tikzpicture}
    \caption{Diffusion-driven purification introduces noise to adversarial images by following the forward diffusion process with a small diffusion timestep \(t^*\) to obtain the diffused images. These images are then denoised through the reverse denoising process to recover the clean images before classification.}
    \label{fig:diffusion_purification}
\end{figure*}
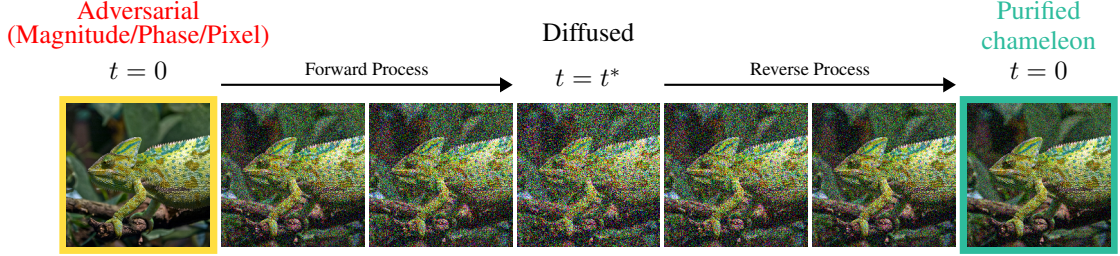

\section{Attack and Defense Mechanisms}

In this section, we provide a detailed description of the attack method and the purification technique used.

\subsection{Frequency-Based Attacks} 

Targeting different image components creates distinct distortion patterns within the image, which also vary depending on the chosen model. Figure~\ref{fig:Frequency_adversarial_examples} illustrates various attack distortion patterns by perturbing the pixel values, phase, amplitude, or a combination of these elements in a clean image (leftmost in Figure~\ref{fig:clean_adv_diff_pure}),  as described below. 

An image \( x \) in the spatial domain can be represented in the frequency domain through the Discrete Fourier Transform (DFT)~\cite{cooley1965algorithm}, which decomposes the image into its frequency components. The Fourier transform of an image \( x \) can be expressed as:

\begin{equation}
\mathcal{F}\{x\} = M \cdot e^{i\phi}, 
\end{equation}

where \( M \) and \( \phi \) represent the magnitude (amplitude) and phase spectra, respectively. Following~\cite{kim2024exploring}, to generate an adversarially perturbed image \( x' \), we apply a combination of multiplicative magnitude perturbation \( \delta_{\text{mag}} \), additive phase perturbation \( \delta_{\text{phase}} \), and additive pixel perturbation \( \delta_{\text{pixel}} \). The perturbed image \( x' \) is thus given by:

\begin{equation}
\tilde{x}' = \mathcal{F}^{-1} \left( \text{clip}_{0,\infty} \left( M \otimes \delta_{\text{mag}} \right) \cdot e^{i(\phi + \delta_{\text{phase}})} \right) + \delta_{\text{pixel}}, 
\end{equation}

where \( \mathcal{F}^{-1} \) denotes the inverse Fourier transform, \( \otimes \) represents element-wise multiplication, and \( \text{clip}_{0,\infty} \) truncates the values of the magnitude spectrum within the range \([0, \infty)\). The image \( \tilde{x}' \) is the intermediate image obtained after the inverse Fourier transform. 

To ensure that the pixel values remain within the valid range \([0, 1]\), the resulting image is clipped as follows:

\begin{equation}
x' = \text{clip}_{0,1} \left( \tilde{x}' \right), 
\end{equation}

The magnitude perturbation \( \delta_{\text{mag}} \) modifies the strength of the frequency components, while the phase perturbation \( \delta_{\text{phase}} \) alters the spatial alignment, resulting in structural changes that may not be visually perceptible. The term \( \delta_{\text{pixel}} \) represents an additive perturbation in the spatial domain.

To maintain real-valued pixel outputs from the inverse Fourier transform, the perturbations \( \delta_{\text{mag}} \) and \( \delta_{\text{phase}} \) are kept symmetric. Adversarial attacks can be performed using a single perturbation \( \delta_{\text{mag}} \), \( \delta_{\text{phase}} \), or \( \delta_{\text{pixel}} \), which we refer to as a magnitude attack, phase attack, and pixel attack, respectively. Additionally, combinations of multiple perturbations are possible, which we also explore in this work.

To obtain the adversarial image \( x' \), the perturbations are optimized via gradient descent. To limit distortion (for imperceptibility)  while maximizing the cross-entropy loss and thereby misleading the classifier, the optimization process minimizes the \( \ell_2 \) difference between the original and perturbed images. The loss function is formulated as: 

\begin{equation}
\mathcal{L} = \lambda \cdot \ell_2(x', x) - \sum_{k} y_k \log f_k(x')  \label{eq:loss_frequency_CE}  
\end{equation}

where \( \lambda \) is a balancing parameter that controls the trade-off between distortion and classification loss. Here, \( f_k(x') \) represents the classifier’s predicted probability for class \( k \), and \( y \) is the one-hot ground truth label.

\subsection{Diffusion-Driven Purification Defense}
Here, we first overview continuous-time diffusion models, followed by the diffusion purification method used.
\vspace{2pt}

\textbf{Continuous-Time Diffusion Models.} Diffusion-based probabilistic generative models operate by progressively corrupting training data with noise and then learning to reverse this corruption, effectively modeling the underlying data distribution~\cite{song2019generative, ho2020denoising, dhariwal2021diffusion, song2021scorebased}. Denoising Diffusion Probabilistic Models (DDPMs) learn to generate data by reversing a Markovian diffusion process that maps complex data distributions to a simple prior, typically an isotropic Gaussian, refining noise into structured samples~\cite{sohl2015deep, ho2020denoising, nichol2021improved}. Instead of applying noise in discrete steps, continuous-time diffusion models use stochastic differential equations (SDEs) to gradually transform data into noise and vice versa~\cite{song2021scorebased}.

Formally, given the unknown data distribution \( p(x) \) from which each data point \( x \in \mathbb{R}^d \) is sampled, the diffusion process gradually transitions \( p(x) \) towards a noise distribution. The forward diffusion process \( \{ x(t) \}_{t \in [0,1]} \), can be described by the following stochastic differential equation:

\begin{equation}
\label{eq:SDE}
dx = h(x, t) dt + g(t) dw,
\end{equation}

where \(x(0):=x \sim p(x) \),  \( h : \mathbb{R}^d \times \mathbb{R} \to \mathbb{R}^d \) represents the drift coefficient, \( g : \mathbb{R} \to \mathbb{R} \) is the diffusion coefficient, and \( w(t) \in \mathbb{R}^n \) is the standard Wiener process  (a.k.a., Brownian motion).  
Under suitable conditions conditions~\cite{anderson1982reverse, song2021scorebased}, the reverse process exists and undo the added noise by solving the reverse-time SDE:

\begin{equation}
\label{eq:Reverse-SDE}
d\hat{x} = \left[ h(\hat{x}, t) - g(t)^2 \nabla_{\hat{x}} \log p_t(\hat{x}) \right] dt + g(t) d\overline{w},
\end{equation}
where \( d\overline{w} \) represents the reverse-time standard Wiener process, and \(\nabla_{x} \log p_t(x)\) denotes the time-dependent score function. Following the conventions of the variance-preserving stochastic differential equation (VP-SDE)~\cite{song2021scorebased}, the drift and diffusion terms are given by

\begin{equation}
h(x; t) = -\frac{1}{2} \beta(t) x, \quad g(t) = \sqrt{\beta(t)},
\end{equation}

with \( \beta(t) \) representing a time-dependent noise scale, which is positive and continuous over the interval \( [0,1] \), such that the state at time \( t \) can be rewritten as:

\begin{equation}
\label{eq:diffused_img}
x(t) = \sqrt{\alpha_t} x(0) + \sqrt{1 - \alpha_t} \epsilon,
\end{equation}

where \(\alpha_t = e^{-\int_0^t \beta(s) ds}\), and \( \epsilon \sim \mathcal{N}(0, I) \).
We denote the reverse-SDE by \( \{ \hat{x}_t \}_{t \in [0,1]} \) from Eq. \ref{eq:Reverse-SDE}, which follows the same distribution as the forward-SDE \( \{ x_t \}_{t \in [0,1]} \) from Eq. \ref{eq:SDE}.

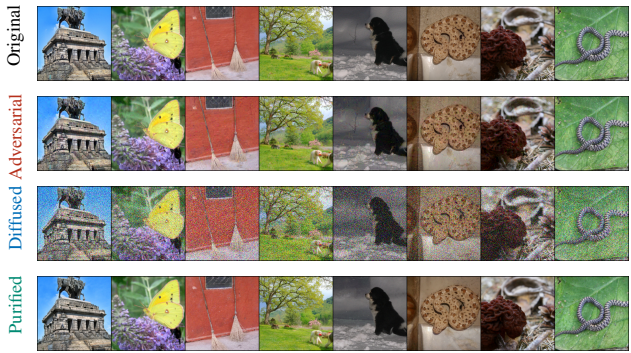
\begin{figure}[h]
    \centering
    \begin{tikzpicture}
        \node (img1) [anchor= west] at (0, 0) {\includegraphics[width=0.45\textwidth]{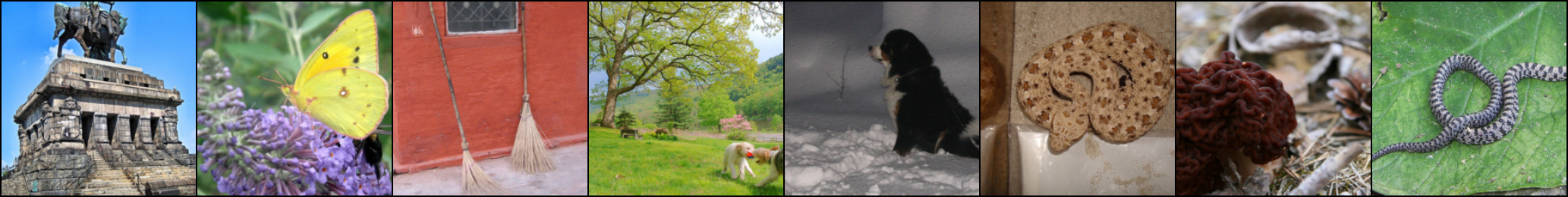}};
        \node[anchor=east, rotate=90, xshift=0.5pt, yshift=5pt] at (img1.north west) {\scriptsize \textcolor{black}{Original}};
        
        \node (img2) [anchor= west] at (0, -1.2) {\includegraphics[width=0.45\textwidth]{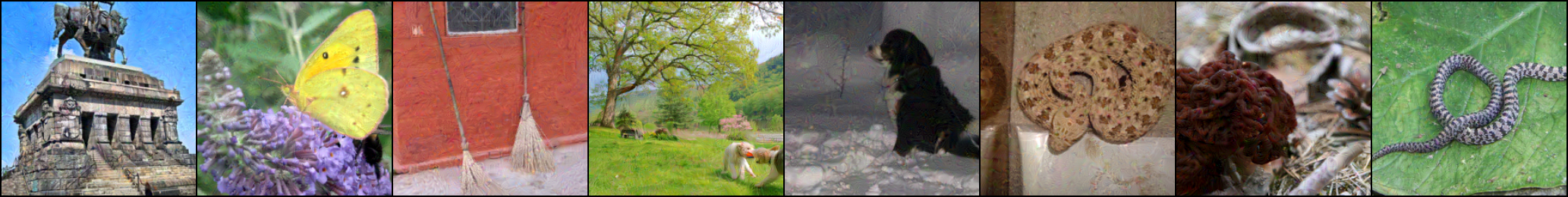}};
        \node[anchor=east, rotate=90, xshift=1pt, yshift=5pt] at (img2.north west) {\scriptsize \textcolor{BrickRed}{Adversarial}};
        
        \node (img3) [anchor= west] at (0, -2.4) {\includegraphics[width=0.45\textwidth]{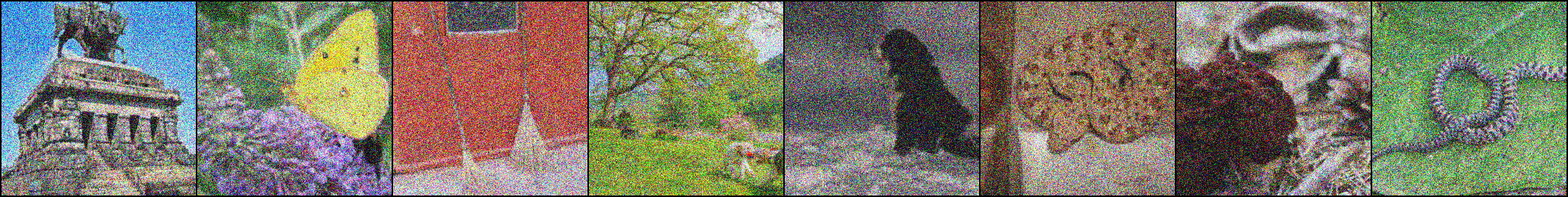}};
        \node[anchor=east, rotate=90, xshift=0.1pt, yshift=5pt] at (img3.north west) {\scriptsize \textcolor{NavyBlue}{Diffused}};
        
        \node (img4) [anchor= west] at (0, -3.6) {\includegraphics[width=0.45 \textwidth]{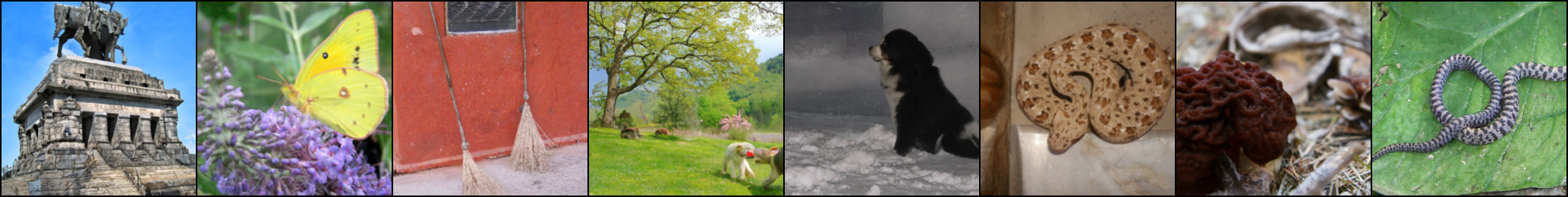}};
        \node[anchor=east, rotate=90, xshift=0.5pt, yshift=5pt] at (img4.north west) {\scriptsize \textcolor{PineGreen}{Purified}};
    \end{tikzpicture}
    \caption{Example of purified images for a pixel attack on the ResNet-50 model with \( t^* = 0.15 \).}
    \label{fig:sample_purified_examples}
\end{figure}

\textbf{Diffusion-Driven Adversarial Purification.} Similar to~\cite{nie2022diffusion}, we employ a two-step adversarial purification method. We begin with an adversarial example \( x' = x(0) \) at time \( t = 0 \) and remove its noise through a diffusion model. The first step involves diffusing the input by solving a forward SDE, as given in Eq.~\ref{eq:SDE}, which gradually adds noise from \( t = 0 \) to \( t = t^* \). Specifically, for a chosen diffusion timestep \( t \in [0, 1] \), the diffused sample is obtained from Eq.~\ref{eq:diffused_img}. Once the adversarial input has been sufficiently diffused, we reverse the process by solving the reverse SDE in Eq.~\ref{eq:Reverse-SDE} from \( t = t^* \) to obtain a purified input \( \hat{x}(0) \). This purified input is subsequently passed to a standard classifier for prediction. This is illustrated in Figure~\ref{fig:diffusion_purification}. Examples of clean, pixel-attacked, diffused, and purified images are shown in Figure~\ref{fig:sample_purified_examples}.

\section{Experimental Setup}

Below, we detail the dataset, training configuration, employed networks, and evaluation metrics, followed by the presentation of results and analysis.

\subsection{Configurations and Metrics}
We aim to explore the effectiveness of adversarial purification against perturbations in the spatial and spectral domains across multiple pre-trained classifier architectures. Specifically, we evaluate ResNet-50~\citep{he2016deep} as a CNN, ViT-B-16~\cite{dosovitskiy2021an} as a transformer model, and Swin-B~\citep{liu2021swin} as a hybrid architecture for a comprehensive evaluation on ImageNet~\cite{deng2009imagenet} dataset.

We minimize the loss in Eq.~\ref{eq:loss_frequency_CE} using the Adam optimizer~\cite{kingma2014adam} with a fixed learning rate of \(5 \times 10^{-3}\) and a weight decay of \(5 \times 10^{-6}\). The optimization is performed for a maximum of 1000 iterations, with a termination condition that halts the process if the loss does not improve over five consecutive iterations. To control the attack strength, we use the value of \( \lambda = 5 \times 10^4 \), which is the default setting employed by \cite{kim2024exploring}.

 For diffusion purification, we follow the experimental setup in~\cite{nie2022diffusion}.  Specifically, we use the adjoint framework for SDEs provided by the TorchSDE library~\cite{li2020scalable,kidger2021neuralsde} for both adversarial purification and gradient evaluation. Both SDEs are solved using the simple Euler-Maruyama method with a fixed step size of \( dt = 10^{-3} \). Additionally, we use the \( 256 \times 256 \) unconditional diffusion checkpoint from the Guided Diffusion library~\cite{guided_diffusion}.

\textbf{Evaluation metrics. } We evaluate models using both standard and robust accuracy. In Tables~\ref{tab:Frequency_results} and~\ref{tab:Frequency_results_timespteps}, "Clean (No Attack)" represents the performance of the pretrained classifier on the entire test set of clean ImageNet data, without any purification. Similarly, "Adversarial" denotes the performance of the model on adversarial examples generated from spectral and spatial attacks.  
The "Purified Clean" refers to the accuracy of the pretrained classifier on clean test samples that have been processed through the diffusion model for purification before classification. This metric allows us to assess the impact of purification on clean accuracy and analyze the trade-off between improved robustness and potential accuracy loss on unperturbed samples.  
The "Adversarial Purified" measures the classifier’s performance on adversarial examples (from spectral and spatial attacks) that are first processed through the diffusion model for purification before classification.  
Following~\cite{nie2022diffusion}, adversarial accuracies are evaluated on a fixed subsets of 512 randomly sampled images. Specifically, we use three subsets, each containing 512 samples, and each experiment is repeated three times. The results are reported as the mean and standard deviation.

We also compare the trade-off between the clean accuracy of the pre-trained model and the clean accuracy achieved through adversarial training and diffusion purification, evaluated on a fixed subset of 512 randomly sampled images from the test set. The adversarially trained model is chosen to be robust against AutoAttack (\(\varepsilon = 4/255\), \(\ell_{\infty}\)) from RobustBench~\cite{croce2020reliable}. Specifically, for ResNet-50~\citep{he2016deep}, we use the adversarially trained model from~\cite{salman2020adversarially}, and for ViT-B-16~\cite{dosovitskiy2021an} and Swin-B~\citep{liu2021swin} , we use the models from~\cite{mo2022adversarial}. For purification, the diffusion timestep is set to \( t^* = 0.15 \). Additionally, to evaluate the effect of diffusion timestep, we consider \( t^* = 0.1 \) and \( t^* = 0.125 \).

\section{Results}

Figure~\ref{fig:Frequency_adversarial_examples} clearly illustrates that different models generate distinct distortion patterns. Moreover, perturbing various image components (pixel, phase, and amplitude) within the same model also leads to differing distortion patterns (both high and low frequency). To better understand how purification handles these diverse distortion patterns, Table~\ref{tab:Frequency_results} presents the performance of diffusion purification across various adversarial perturbations applied to different network architectures.

The pixel attack is most effective on the ResNet-50 model, and purification effectively restores the adversarial sample, enhancing its robustness. Similarly, for spectral attacks, diffusion purification is highly effective at removing perturbations. On average, across both spatial and spectral perturbations, the robustness improves by up to \(7.72 \times 10^2\%\) compared to the pre-trained model. In terms of clean accuracy, there is an average drop of 8.62\% across all perturbations compared to the pre-trained model, due to the clean input passing through adversarial purification. This drop is typically not reported in previous studies~\cite{nie2022diffusion}; however, in real-world scenarios, we do not know in advance which samples are clean and which are adversarially crafted. Although clean accuracy decreases, as shown in Table~\ref{tab:model_performance_Robustbench}, purification still results in a smaller trade-off, and offering better robustness compared to adversarial training.
Similarly, for the ViT and Swin-B models, adversarial perturbations degrade performance in both spectral and spatial domains, but purification results in significant benefits, even more so than for the CNN model. In terms of robustness, although there is a clean accuracy drop of 5.60\% for ViT and 6.79\% for Swin-B on average across all perturbations due to purifying clean samples, the robustness gain is as high as \(8.20 \times 10^3\%\) for ViT and \(7.73 \times 10^5\%\) for Swin-B compared to the pre-trained model after purification.

Table \ref{tab:model_performance_Robustbench} presents the trade-off between the clean accuracy of the pre-trained models and the clean accuracy achieved through adversarial training~\cite{tsipras2018robustness, zhang2019theoretically} and diffusion purification. 
While both adversarial training and diffusion purification reduce the clean accuracy of the pre-trained model across all models, diffusion purification incurs less accuracy loss and achieves superior robustness compared to adversarial training~\cite{nie2022diffusion}.

Table~\ref{tab:Frequency_robustenss_autoattack_models} compares the performance of robust adversarially trained models against unseen frequency perturbations and purification. On average, across all perturbations, the performance gain with purification is 66.49\% for the ResNet-50 model, compared to adversarial training, and 15.97\% for the ViT model.

Table~\ref{tab:Frequency_results_timespteps} compares the impact of diffusion timestep \(t^*\). For the ResNet-50 model, across all perturbations, a smaller timestep \(t^* = 0.1\) results in better clean accuracy, while larger diffusion timesteps enhance adversarial robustness. In the case of the ViT-b model, the timestep has a relatively minor effect, with similar performance observed across various timesteps \(t^* = 0.1\), \(t^* = 0.125\), and \(t^* = 0.15\) (as shown in Table~\ref{tab:Frequency_results}). For the Swin-b model, a smaller \(t^*\) achieves better adversarial gains.

\section{Conclusions}

In this study, we employed the Diffusion model to evaluate the effectiveness of purification in countering diverse adversarial perturbations across both spectral and spatial domains. Our findings consistently demonstrate the model's robustness across all tested scenarios, emphasizing that purification is highly effective in restoring robustness against unseen and varied adversarial perturbations while incurring minimal trade-offs.

\begin{table}[h]
    \centering
    \renewcommand{\arraystretch}{1.1} 
\setlength{\tabcolsep}{5pt} 
    \begin{tabular}{lccc}
        \toprule
        Model & \shortstack{Clean \\ (Pre-trained)} & \shortstack{Clean \\ (Robust)} & \shortstack{Clean \\ (Purified)} \\
        \midrule
        \shortstack{ResNet-50 \\ \scriptsize \textcolor{MidnightBlue}{(Salman et al., 2020)~\cite{salman2020adversarially}} }
        & 75.78 & 63.86 & 70.89 \\
        \shortstack{ViT-B \\ \scriptsize \textcolor{MidnightBlue}{(Mo et al., 2022)~\cite{mo2022adversarial}} }
        & 82.61 & 67.77 & 77.73 \\
        \shortstack{Swin-B \\ \textcolor{MidnightBlue}{\scriptsize(Mo et al., 2022)~\cite{mo2022adversarial}}} 
        & 84.76 & 74.8 & 81.05 \\
        \bottomrule
    \end{tabular}
    \caption{Clean accuracy trade-off between purification and adversarially trained models on ImageNet under AutoAttack (\(\varepsilon = 4/255\), \(\ell_{\infty}\)). Adversarially trained models are sourced from RobustBench~\cite{croce2020reliable}, while purification is applied with a diffusion timestep of \( t^* = 0.15 \). Clean (Pre-trained), Clean (Robust), and Clean (Purified) refer to the clean accuracy of the pre-trained model, adversarially trained model, and purified AutoAttack~\cite{croce2020reliable} examples, respectively. Adversarial purification leads to smaller reductions in clean accuracy while achieving greater improvements in robustness (Tables~\ref{tab:Frequency_results} and~\ref{tab:Frequency_robustenss_autoattack_models} and~\cite{nie2022diffusion}).}
    \label{tab:model_performance_Robustbench}
\end{table}


\begin{table}[hbt!] 
\centering
\renewcommand{\arraystretch}{1.1} 
\setlength{\tabcolsep}{5pt} 
\begin{tabular}{llcc}
\toprule
\multicolumn{2}{c}{\multirow{2}{*}{}} & \multicolumn{2}{c}{Accuracy} \\
\cmidrule{3-4}
\multicolumn{2}{c}{\multirow{2}{*}{}} & Adversarial & Purified Adversarial \\
\midrule
\multirow{6}{*}{\rotatebox{90}{\shortstack{ResNet-50 \\ \scriptsize \textcolor{MidnightBlue}{(Salman et al., 2020)~\cite{salman2020adversarially}}}}} 
& Pixel 
& 29.49
&  63.86
\\
& Mag 
&  48.05
&  68.75
\\
& Phase 
& 49.61 
&  67.77
\\
& Phase+Mag 
&  41.02
&  66.61
\\
& All 
&  29.49
&  64.06
\\
\arrayrulecolor{black!20}\cmidrule{2-4}
& Mean 
&  39.53
&  66.21
\\
\arrayrulecolor{black}\cmidrule{2-4}

\multirow{6}{*}{\rotatebox{90}{\shortstack{ViT-B-16 \\ \textcolor{MidnightBlue}{\scriptsize(Mo et al., 2022)~\cite{mo2022adversarial}} } }} 
& Pixel 
& 63.28 
&  74.02
\\
& Mag 
& 65.42 
& 76.37 
\\
& Phase 
&  66.01
&  75.19
\\
& Phase+Mag 
&  63.67
&  73.44
\\
& All 
&  63.08
&  73.82
\\
\arrayrulecolor{black!20}\cmidrule{2-4}
& Mean 
&  64.29
&  74.56
\\
\arrayrulecolor{black}\cmidrule{1-4}
\end{tabular}
\caption{Comparison of the robustness of purification and adversarially trained models against frequency-based perturbations. Adversarial purification shows superior robustness in countering frequency attacks.}
\label{tab:Frequency_robustenss_autoattack_models}
\end{table}

\begin{table*}[hbt!] 
\centering
\renewcommand{\arraystretch}{1.2} 
\setlength{\tabcolsep}{4pt} 
\begin{tabular}{lllcccc}
\toprule
\multicolumn{3}{c}{\multirow{2}{*}{}} & \multicolumn{4}{c}{Accuracy} \\
\cmidrule{4-7}
\multicolumn{3}{c}{} &  \text{Clean (No Attack)} &  \text{Adversarial} & \text{Purified Clean} &  \text{Purified Adversarial} \\
\midrule
&
\multirow{6}{*}{\rotatebox{90}{ResNet-50}}
& pixel 
& 76.69\scriptsize\textcolor{gray!95}{$\pm${1.12}}
& 1.17\scriptsize\textcolor{gray!95}{$\pm${0.35}}
& 70.08\scriptsize\textcolor{gray!95}{$\pm${0.96}}
& 64.29\scriptsize\textcolor{gray!95}{$\pm${1.48}}
\\
&& mag 
& 76.69\scriptsize\textcolor{gray!95}{$\pm${1.12}}
& 12.96\scriptsize\textcolor{gray!95}{$\pm${1.62}}
& 70.08\scriptsize\textcolor{gray!95}{$\pm${0.96}}
& 67.06\scriptsize\textcolor{gray!95}{$\pm${1.23}}
\\
&& phase 
& 76.69\scriptsize\textcolor{gray!95}{$\pm${1.12}}
& 12.73\scriptsize\textcolor{gray!95}{$\pm${1.54}}
& 70.12\scriptsize\textcolor{gray!95}{$\pm${1.00}}
& 67.02\scriptsize\textcolor{gray!95}{$\pm${1.70}}
\\
&& phase+mag 
& 76.69\scriptsize\textcolor{gray!95}{$\pm${1.12}}
& 9.70\scriptsize\textcolor{gray!95}{$\pm${1.47}}
& 70.05\scriptsize\textcolor{gray!95}{$\pm${0.98}}
& 65.27\scriptsize\textcolor{gray!95}{$\pm${0.92}}
\\
&& all 
& 76.69\scriptsize\textcolor{gray!95}{$\pm${1.12}}
& 1.04\scriptsize\textcolor{gray!95}{$\pm${0.27}}
& 70.08\scriptsize\textcolor{gray!95}{$\pm${0.96}}
& 64.26\scriptsize\textcolor{gray!95}{$\pm${1.32}}
\\
\arrayrulecolor{black!20}\cmidrule{3-7}
&& mean 
& 76.69\scriptsize\textcolor{gray!95}{$\pm${1.04}}
& 7.52\scriptsize\textcolor{gray!95}{$\pm${5.57}}
& 70.08\scriptsize\textcolor{gray!95}{$\pm${0.90}}
& 65.58\scriptsize\textcolor{gray!95}{$\pm${1.79}}
\\
\arrayrulecolor{black}\cmidrule{2-7}
&\multirow{6}{*}{\rotatebox{90}{ ViT-B-16}}
& pixel 
& 80.47\scriptsize\textcolor{gray!95}{$\pm${1.68}}
& 0.16\scriptsize\textcolor{gray!95}{$\pm${0.15}}
& 75.98\scriptsize\textcolor{gray!95}{$\pm${1.22}}
& 74.54\scriptsize\textcolor{gray!95}{$\pm${1.86}}
\\
&& mag 
& 80.47\scriptsize\textcolor{gray!95}{$\pm${1.68}}
& 1.89\scriptsize\textcolor{gray!95}{$\pm${0.34}}
& 75.98\scriptsize\textcolor{gray!95}{$\pm${1.22}}
& 74.71\scriptsize\textcolor{gray!95}{$\pm${1.69}}
\\
&& phase 
& 80.47\scriptsize\textcolor{gray!95}{$\pm${1.68}}
& 1.92\scriptsize\textcolor{gray!95}{$\pm${0.23}}
& 75.94\scriptsize\textcolor{gray!95}{$\pm${1.21}}
& 75.03\scriptsize\textcolor{gray!95}{$\pm${1.65}}
\\
&& phase+mag 
& 80.47\scriptsize\textcolor{gray!95}{$\pm${1.68}}
& 0.46\scriptsize\textcolor{gray!95}{$\pm${0.24}}
& 75.98\scriptsize\textcolor{gray!95}{$\pm${1.22}}
& 74.74\scriptsize\textcolor{gray!95}{$\pm${1.59}}
\\
&& all 
& 80.47\scriptsize\textcolor{gray!95}{$\pm${1.68}}
& 0.07\scriptsize\textcolor{gray!95}{$\pm${0.10}}
& 75.94\scriptsize\textcolor{gray!95}{$\pm${1.21}}
& 74.67\scriptsize\textcolor{gray!95}{$\pm${1.76}}
\\
\arrayrulecolor{black!20}\cmidrule{3-7}
&& mean 
& 80.47\scriptsize\textcolor{gray!95}{$\pm${1.56}}
& 0.90\scriptsize\textcolor{gray!95}{$\pm${0.87}}
& 75.96\scriptsize\textcolor{gray!95}{$\pm${1.13}}
& 74.74\scriptsize\textcolor{gray!95}{$\pm${1.60}}
\\
\arrayrulecolor{black}\cmidrule{2-7}
&\multirow{6}{*}{\rotatebox{90}{ Swin-B}}
& pixel 
& 84.57\scriptsize\textcolor{gray!95}{$\pm${0.46}}
& 0.07\scriptsize\textcolor{gray!95}{$\pm${0.10}}
& 78.84\scriptsize\textcolor{gray!95}{$\pm${1.38}}
& 77.31\scriptsize\textcolor{gray!95}{$\pm${1.04}}
\\
&& mag 
& 84.57\scriptsize\textcolor{gray!95}{$\pm${0.46}}
& 0.00\scriptsize\textcolor{gray!95}{$\pm${0.00}}
& 78.81\scriptsize\textcolor{gray!95}{$\pm${1.36}}
& 77.34\scriptsize\textcolor{gray!95}{$\pm${0.95}}
\\
&& phase 
& 84.57\scriptsize\textcolor{gray!95}{$\pm${0.46}}
& 0.00\scriptsize\textcolor{gray!95}{$\pm${0.00}}
& 78.84\scriptsize\textcolor{gray!95}{$\pm${1.38}}
& 77.41\scriptsize\textcolor{gray!95}{$\pm${0.77}}
\\
&& phase+mag 
& 84.57\scriptsize\textcolor{gray!95}{$\pm${0.46}}
& 0.00\scriptsize\textcolor{gray!95}{$\pm${0.00}}
& 78.84\scriptsize\textcolor{gray!95}{$\pm${1.38}}
& 77.41\scriptsize\textcolor{gray!95}{$\pm${0.67}}
\\
&& all 
& 84.57\scriptsize\textcolor{gray!95}{$\pm${0.46}}
& 0.00\scriptsize\textcolor{gray!95}{$\pm${0.00}}
& 78.81\scriptsize\textcolor{gray!95}{$\pm${1.36}}
& 77.25\scriptsize\textcolor{gray!95}{$\pm${1.02}}
\\
\arrayrulecolor{black!20}\cmidrule{3-7}
&& mean 
& 84.57\scriptsize\textcolor{gray!95}{$\pm${0.43}}
& 0.01\scriptsize\textcolor{gray!95}{$\pm${0.05}}
& 78.83\scriptsize\textcolor{gray!95}{$\pm${1.28}}
& 77.34\scriptsize\textcolor{gray!95}{$\pm${0.84}}
\\
\arrayrulecolor{black}\cmidrule{2-7}
\end{tabular}
\caption{Performance of diffusion purification on various types of adversarial perturbations across different networks on ImageNet. "Clean (No Attack)" and "Adversarial" denote classifier performance on clean and adversarial examples without purification. "Purified Clean" and "Adversarial Purified" assess accuracy on clean and adversarial samples after purification using diffusion timesteps \( t^* = 0.15 \). Purification enhances robustness across various perturbations and models, with notable improvements in both pixel and spectral attacks. While clean accuracy drops slightly due to the purification process, robustness gains are substantial, particularly for ViT and Swin-B models.}
\label{tab:Frequency_results}
\end{table*}
%
%
%
%
\begin{table*}[hbt!] 
\centering
\renewcommand{\arraystretch}{1.1} 
\setlength{\tabcolsep}{4pt} 
\begin{tabular}{lllcccccc}
\toprule
\multicolumn{3}{c}{\multirow{2}{*}{}} & \multicolumn{6}{c}{Accuracy} \\
\cmidrule{4-9}
\multicolumn{3}{c}{\multirow{3}{*}{}} & \multicolumn{2}{c}{} & \multicolumn{2}{c}{Purified Clean} & \multicolumn{2}{c}{Purified Adversarial} \\
\cmidrule(lr){6-7} \cmidrule(lr){8-9}
\multicolumn{3}{c}{} &  \text{Clean (No Attack)} &  \text{Adversarial} & \text{\(t^*=0.125\)} & \text{\(t^*=0.1\)} & \text{\(t^*=0.125\)} & \text{\(t^*=0.1\)}  \\
\midrule
  &
\multirow{6}{*}{\rotatebox{90}{ResNet-50}}
& pixel 
& 78.13  
& 1.56  
& 72.66  
& 73.05  
& 65.43  
& 61.72   \\
&& mag 
& 78.13  
& 10.94  
& 72.66  
& 73.05  
& 66.02  
& 65.63   \\
&& phase 
& 78.13  
& 12.89  
& 72.66  
& 73.05  
& 67.38  
& 68.16   \\
&& phase+mag 
& 78.13  
& 7.62  
& 72.66  
& 73.05  
& 63.48  
& 60.94   \\
&& all 
& 78.13  
& 1.95  
& 72.66  
& 73.05  
& 64.65  
& 60.74   \\
\arrayrulecolor{black!20}\cmidrule{3-9}
&& mean 
& 78.13  
& 6.99  
& 72.66  
& 73.05  
& 65.39  
& 63.44   \\
\arrayrulecolor{black}\cmidrule{2-9}
&\multirow{6}{*}{\rotatebox{90}{ViT-B-16}}
& pixel 
& 79.10  
& 0.00  
& 76.56  
& 75.78  
& 75.00  
& 74.61   \\
&& mag 
& 79.10  
& 1.76  
& 76.56  
& 75.78  
& 74.22  
& 73.83   \\
&& phase 
& 79.10  
& 2.34  
& 76.56  
& 75.78  
& 75.00  
& 74.41   \\
&& phase+mag 
& 79.10  
& 0.20  
& 76.56  
& 75.78  
& 75.20  
& 73.63   \\
&& all 
& 79.10  
& 0.00  
& 76.56  
& 75.78  
& 74.80  
& 74.80   \\
\arrayrulecolor{black!20}\cmidrule{3-9}
&& mean 
& 79.10  
& 0.86  
& 76.56  
& 75.78  
& 74.84  
& 74.26   \\
\arrayrulecolor{black}\cmidrule{2-9}
&\multirow{6}{*}{\rotatebox{90}{Swin-B}}
& pixel 
& 84.96  
& 0.00  
& 78.32  
& 80.27  
& 78.91  
& 78.91   \\
&& mag 
& 84.96  
& 0.00  
& 78.32  
& 80.27  
& 78.91  
& 78.71   \\
&& phase 
& 84.96  
& 0.00  
& 78.32  
& 80.27  
& 79.10  
& 80.08   \\
&& phase+mag 
& 84.96  
& 0.20  
& 78.32  
& 80.27  
& 77.93  
& 78.71   \\
&& all 
& 84.96  
& 0.00  
& 78.32  
& 80.27  
& 78.71  
& 80.66   \\
\arrayrulecolor{black!20}\cmidrule{3-9}
&& mean 
& 84.96  
& 0.02  
& 78.32  
& 80.27  
& 78.71  
& 79.41   \\
\arrayrulecolor{black}\cmidrule{2-9}
\end{tabular}
\caption{The impact of diffusion timestep \( t^* \) on the performance of diffusion purification across various perturbations and networks on ImageNet. Overall, larger diffusion timesteps enhance robustness, while smaller timesteps yield better clean accuracy, although they reduce the effectiveness of adversarial purification. }
\label{tab:Frequency_results_timespteps}
\end{table*}

%
%

\section*{Acknowledgments}
This publication has emanated from research conducted with the financial support of Taighde Éireann – Research Ireland under Grant No. 18/CRT/6223.

{
    \small
    \bibliographystyle{ieeenat_fullname}
    \bibliography{main}
}


\end{document}